\documentclass[acmsmall,nonacm]{acmart}

\usepackage{xcolor}
\usepackage{booktabs}
\definecolor{gray1}{HTML}{e4e4e4}
\definecolor{blue1}{HTML}{e0ffff}

\usepackage{graphicx}

\usepackage[most,breakable]{tcolorbox}
\tcbset{
  boxsep=0mm,boxrule=0.5pt
}

\usepackage{subcaption}

\makeatletter
\def\@journalNameShort{}
\def\@journalName{}
\makeatother

\newcommand{\nterm}[1]{\textbf{\small \textsf{#1}}}
\newcommand{\kw}[1]{\textcolor{blue}{\textsf{\small #1}}}
\newcommand{\smallsf}[1]{\text{\small\textsf{#1}}}

\setcopyright{none}
\renewcommand\footnotetextcopyrightpermission[1]{}

\usepackage{enumitem}
\setlist[itemize]{leftmargin=1.5em, labelsep=0.5em}
\setlist[enumerate]{leftmargin=1.8em, labelsep=0.5em}

\begin{document}

\title{Social, Legal, Ethical, Empathetic and Cultural Norm Operationalisation for AI Agents}

\author{Radu Calinescu}
\affiliation{%
  \institution{
  University of York}
  \city{York}
  \country{UK}
}
\email{radu.calinescu@york.ac.uk}
\author{Ana Cavalcanti}
\affiliation{%
  \institution{
  University of York}
  \city{York}
  \country{UK}
}
\email{ana.cavalcanti@york.ac.uk}
\author{Marsha Chechik}
\affiliation{%
  \institution{
  University of Toronto}
  \city{Toronto}
  \country{Canada}
}
\email{chechik@cs.toronto.edu}
\author{Lina Marsso}
\affiliation{%
  \institution{
  Polytechnique Montr\'{e}al}
  \city{Montr\'{e}al}
  \country{Canada}
}
\email{lina.marsso@polymtl.ca}
\author{Beverley Townsend}
\affiliation{%
  \institution{
  University of York}
  \city{York}
  \country{UK}
}
\email{bev.townsend@york.ac.uk}

\begin{abstract}
    As AI agents are increasingly used in high-stakes domains like healthcare and law enforcement, aligning their behaviour with social, legal, ethical, empathetic, and cultural (SLEEC) norms has become a critical engineering challenge. While international frameworks have established high-level normative principles for AI, a significant gap remains in translating these abstract principles into concrete, verifiable requirements. To address this gap, we propose a systematic SLEEC-norm operationalisation process for determining, validating, implementing, and verifying normative requirements. Furthermore, we survey the landscape of methods and tools supporting this process, and identify key remaining challenges and research avenues for addressing them.  We thus establish a framework---and define a research and policy agenda---for developing AI agents that are not only functionally useful but also demonstrably aligned with human norms and values.
\end{abstract}

\maketitle

\section{Introduction}

Artificial intelligence (AI) agents are being developed at pace for use in  domains such as healthcare, transportation, law enforcement, and education. These agents operate with limited human oversight in pursuit of high-level goals, which they aim to achieve through sophisticated sequences of agent-selected actions. They are increasingly entrusted with making life-changing decisions~\cite{duan2019artificial,harish2021artificial}, or with controlling critical systems like self-driving cars~\cite{hong2021ai} and assistive-care robots~\cite{worth2024robots}. 

These high-stakes tasks have profound normative implications of a social, legal, ethical, empathetic and cultural (SLEEC) nature.  AI agent actions may even entail prioritising one human value over another, such as favouring privacy over accuracy, or prioritising the users' well-being over their autonomy. To navigate these complexities, we need a comprehensive understanding of the SLEEC norms that reflect the relevant societal and individual values of AI-agent stakeholders. Importantly, we need methods for engineering AI agents that comply with such SLEEC norms. For example, developed with these methods, an AI-enabled care robot will know that, under normal circumstances, the user should be asked for consent before a human operator is called for extra help. Similarly, a tele-health AI chatbot will adapt the content, tone, and volume of its communication to the needs of users, as defined by their cultural backgrounds and possible disabilities.

This level of SLEEC sensitivity and responsiveness in AI agents needs to draw on a plural set of non-exhaustive norms that encode human values derived from rights-based instruments, policy frameworks, standards, professional codes of conduct and best practice, protocols, ethical theory and guidelines, and other sources of social and cultural values. To enable this, a significant corpus of work has been undertaken by multiple organizations, including the OECD~\cite{oecd-2024}, UNESCO~\cite{unesco-2021}, IEEE~\cite{ieee-2021}, ICO/IEC~\cite{iso-iec-2022} and the BSI~\cite{BS-2022}, among others. Nevertheless, normative principles at a high level of abstraction only state that an AI agent should observe human privacy, well-being, autonomy, etc. 
Their high-level framing is not and cannot be specialised for the particularities inherent to contexts and applications, and it tells us little about how to apply and enforce these principles, their contextual appropriateness, and their \emph{operationalisation} for a concrete AI agent. What we mean by SLEEC norm operationalisation is \emph{the complex process of identifying relevant social, legal, ethical, empathetic and cultural norms, and translating them into unambiguous, concrete requirements that are then validated, implemented by an AI agent under development, and verified prior to deploying that agent}.

Addressing the critical question of SLEEC norm operationalisation for concrete AI agents requires reaching down to individual cases and applying norms practically in real-world contexts. For high-level norms to be useful, we need to ground them in evaluative standards specified as \emph{normative requirements} through a systematic process of SLEEC rule elicitation~\cite{townsend2022pluralistic}, validation, and refinement~\cite{feng2024analyzing,feng2024normative,feng2023towards,getir2023specification,Troquard2024}. Furthermore, the AI agent for which a SLEEC ruleset is obtained using this process must be \emph{verified} to ensure its SLEEC-rule compliance~\cite{yaman2023specification,getir2023specification,yaman2024toolkit}. Finally, if on-the-fly code synthesis is used for enhanced autonomy, these rules need to be considered during deployment as well~\cite{sleec-adapt-2025}. Adopting such a SLEEC-norm operationalisation approach is the only way to ensure that AI agent actions in response to stimuli are modelled on those that a human moral agent following the applicable norms would undertake in response to a similar body of information. 

Traditional requirements and systems engineering methods are ill-equipped for this task. They lack the mechanisms to effectively engage the broad range of required technical and non-technical stakeholders, to operate at the high abstraction level of normative principles, to deal with the complex and often subtle conflicts inherent in normative requirements, and to provide SLEEC requirement well-formedness and AI agent compliance guarantees.

This paper explores how recent research advances in SLEEC norm elicitation, validation, and verification~\cite{townsend2022pluralistic,feng2024analyzing,feng2024normative,feng2023towards,getir2023specification,yaman2023specification,yaman2024toolkit,sleec-adapt-2025,townsend2025normative,kleijwegt2025tool} contribute to addressing this gap, and discusses remaining challenges for the systematic operationalisation of normative requirements for AI agents. To that end, we  propose a \emph{SLEEC-norm operationalisation process}. We summarise the  stages and activities of this process, and overview the methods and tools already available to carry out some of these activities. Looking ahead, we then discuss remaining challenges that must be addressed to ensure efficient and effective SLEEC norm operationalisation for the rapidly expanding range of AI technologies and applications, outlining initial steps and avenues to tackle these challenges.

\section{SLEEC Norm Operationalisation Process}

The operationalisation of SLEEC norms requires dedicated activities throughout the AI agent development lifecycle. Many of these activities are often uncommon for other types of AI agent requirements, or pose challenges not encountered for conventional engineered systems. 

\begin{figure}
\centering
\includegraphics[width=\linewidth]{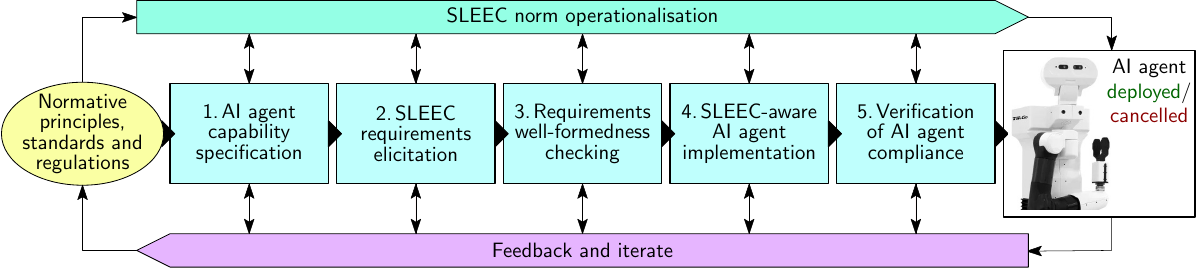}
    \caption{SLEEC norm operationalisation process. Successful completion of all stages is necessary for the deployment of a SLEEC-norm-compliant AI agent; failure at any point precludes the agent's deployment.}
    \Description{Diagram of the SLEEC norm operationalisation process.}
    \label{fig:process}

    \vspace*{-2mm}
\end{figure}

Our proposed process, illustrated in Figure~\ref{fig:process}, consists of five stages. Stages~1 and~2 involve determining the functional capabilities of the AI agent, and eliciting its SLEEC requirements, and are carried out as part of the planning phase of the agent's development lifecycle. Stage~3 encompasses the formal well-formedness checking of the SLEEC requirements, to identify and resolve issues such as conflicts and redundancies; this stage is part of the analysis phase of the lifecycle. Stage~4 translates validated SLEEC requirements into implementable agent abstractions, learning-time data schemas, and runtime guardrails, enabling both enforced compliance at deployment, and controlled adaptation when SLEEC requirements evolve. 
Stage~5, potentially extending across the design, implementation and testing phases of development, comprises the verification of the agent's compliance with its SLEEC requirements. Successful verification leads to the deployment of the AI agent. If verification or any earlier-stage activity fails, the SLEEC norm operationalisation process is halted, potentially precluding the development or deployment of the AI agent. 

\begin{figure}
\centering
\includegraphics[width=0.49\linewidth,trim=9.5cm 0 10cm 0,clip]{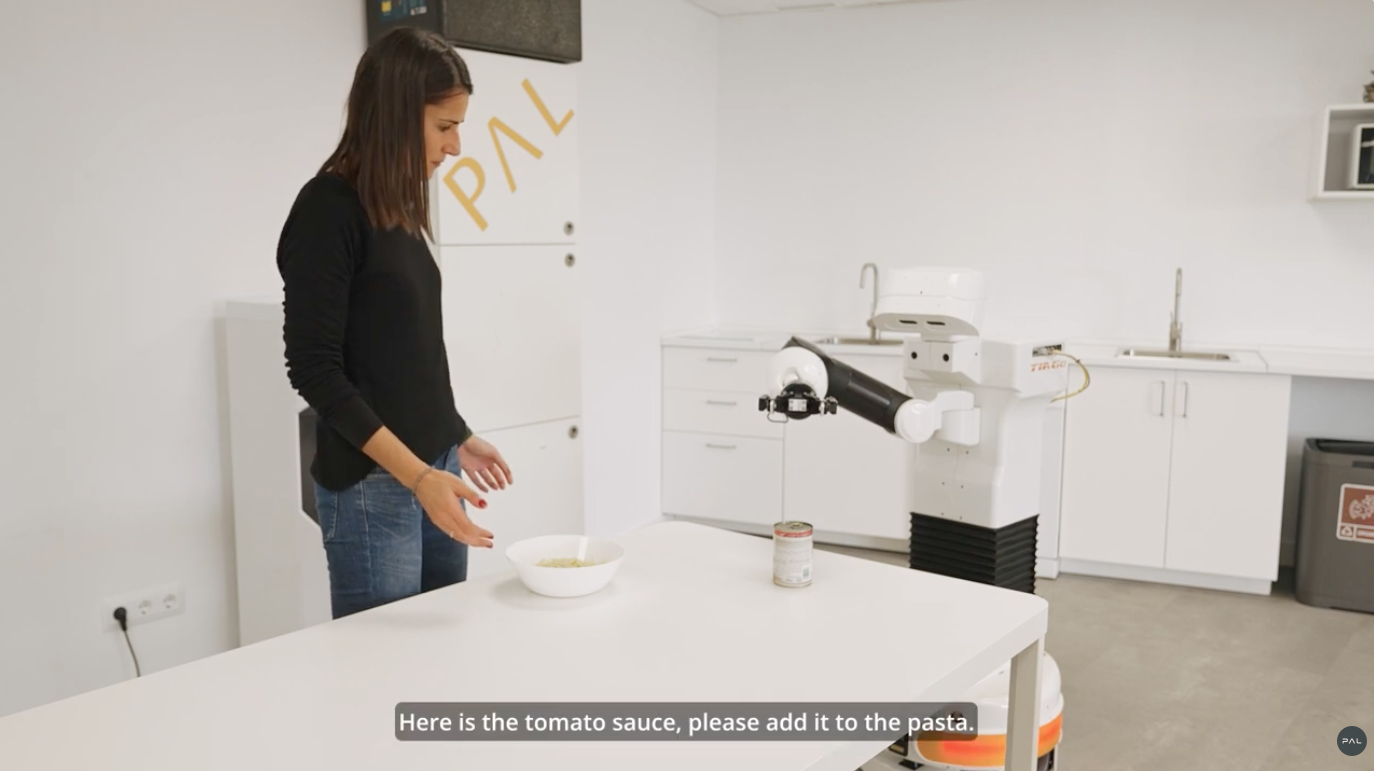}\hspace*{1.5mm}
\includegraphics[width=0.49\linewidth,trim=9.5cm 0 10cm 0,clip]{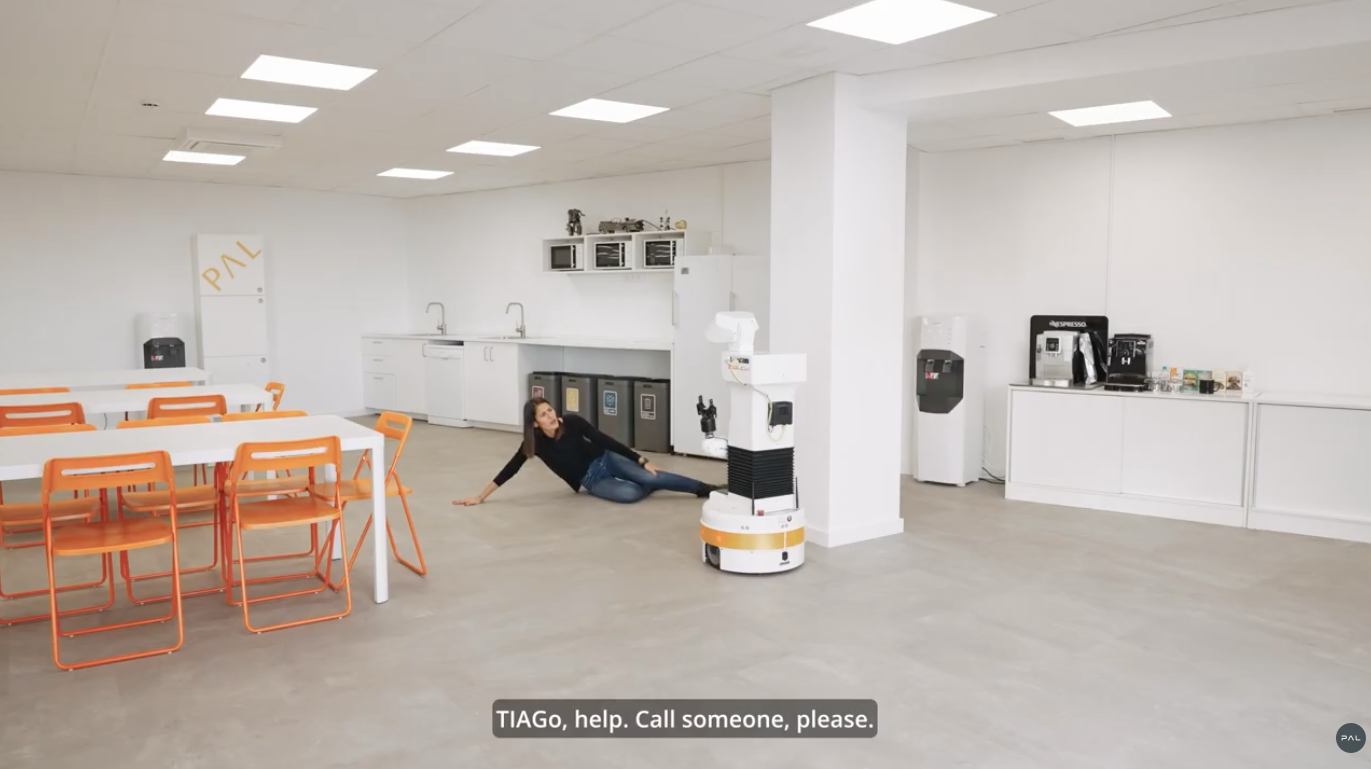}

\vspace*{-1mm}
    \caption{Stills from the 
    ALMI project~\cite{stefanakos-2026} demonstration video at \url{https://youtu.be/VhfQmJe4IPc}. The images show the robot providing user assistance during a cooking task 
    and after a simulated user fall (right).}
    \Description{Stills from the ALMI project demonstration video showing the TIAGo robot used by the project helping the user with the preparation of a pasta meal (left) and providing assistance after to the user who had a fall (right).}
    \label{fig:almi}

    \vspace*{-3mm}
\end{figure}

As indicated in Figure~\ref{fig:process}, the process is inherently iterative: feedback from the activities of a stage may trigger a return to an earlier stage to fine-tune that stage's output. We detail each stage of our proposed process in turn below, using the embodied-AI agent shown in Figure~\ref{fig:almi} as a running example. Prototyped by the \emph{Ambient Assisted Living for Long-term Monitoring and Interaction} (ALMI) project~\cite{almi-2023,stefanakos-2026}, this agent integrates a suite of AI components (covering perception, speech recognition, reasoning, planning, etc.) with a PAL Robotics TIAGo mobile manipulator robot tasked with providing support to a home user with mild cognitive and physical impairments.

\smallskip
\noindent
\textbf{Process input: normative principles, standards and regulations.} The proliferation of AI agents with functionality that is novel or entirely unseen means that developers, operators, users and other stakeholders often lack prior experience to comprehensively identify SLEEC implications and elicit corresponding normative requirements for these agents. Our process leverages the guiding principles, standards and regulations delivered by the sustained global AI governance efforts within recent years. This encompasses AI normative frameworks developed for both general-purpose AI and domain-specific contexts (e.g., healthcare, automotive and finance). 

The existence and continued refinement of these frameworks are essential prerequisites for systematically operationalising the SLEEC requirements of AI agents. 
Crucially, this dependence is reciprocal:~while our methodology is enabled by existing AI governance, its systematic application to concrete AI agents is designed to identify practical gaps, loopholes, and drawbacks within the frameworks themselves, thereby generating actionable insights for their improvement. 

\vspace*{-1mm}
\begin{tcolorbox}[breakable,colback=blue1,colframe=black]
\noindent
{\small 
\textbf{ALMI SLEEC norm operationalisation inputs.} For the ALMI assistive-care robot, SLEEC norm operationalisation should be informed by a hierarchy of normative frameworks. At a global level, these include generic ethics frameworks such as the UNESCO Recommendation on the Ethics of Artificial Intelligence~\cite{unesco-2021} and the OECD AI Principles~\cite{oecd-2024}, which cover principles such as dignity, fairness, and transparency for all applications of AI. Further guidance is provided by international 
standards, including ISO/IEC and IEEE standards supportive of AI quality, transparency, risk management, safety, and ethics in autonomous and intelligent systems. At a regional and domestic level, the normative landscape is shaped by binding legal instruments, notably the EU AI Act~\cite{act2024eu} and the GDPR~\cite{gdpr}---both mandatory since ALMI is intended for use within the European Union. These instruments impose legally enforceable obligations relating to risk classification, safety, transparency, and data protection. In parallel, sectoral 
governance provides normative direction. In the United Kingdom, for example, this includes adherence to a principles-based approach to AI governance that involves layering sectoral laws (e.g., healthcare, consumer, and data protection) onto AI systems like care robots. This is complemented by guidance from relevant regulatory and standard-setting bodies, such as the British Standards Institute, the Information Commissioner's Office, and the Care Quality Commission.  
All these frameworks collectively contribute to defining the normative landscape for the robot's operation.
} 
\end{tcolorbox}

\noindent
\textbf{Stage~1.~AI agent capability specification.} This stage 
encompasses the specification of the high-level functional \emph{capabilities} that determine how  the agent can perceive or affect its environment. Technologically, these capabilities arise from the use of sensors and actuators of a robot, access to communication channels, third-party data access, AI-enabled APIs, and so forth.
The AI agent capabilities considered are those that influence two crucial aspects of SLEEC norm operationalisation:
\begin{enumerate}
    \item {\em Capabilities that determine which SLEEC norms are relevant for an AI agent.} For instance, the care robot from our earlier example may be equipped with a camera and AI-enabled embedded software required to recognise residents at a care home, or to determine whether a vulnerable user has fallen. 
    However, the presence of a camera immediately raises potential normative issues regarding the privacy of the user and any bystanders. 
    \item {\em Capabilities needed to satisfy SLEEC norms}. For example, if the care robot has the capability to identify when a patient is unwell, the `duty of care' principle means that the robot should also possess the capability to enable assistance when that happens, e.g., by leveraging a communication channel to alert support-care personnel. 
\end{enumerate}
Determining the agent capabilities relevant to these two aspects of the operationalisation requires an understanding of the mission of the robot, its prospective robotic platform, and relevant normative frameworks. The output of this stage is a list of informally specified AI-agent capabilities, accompanied by use cases that explain these capabilities to the agent's non-technical stakeholders.

\begin{tcolorbox}[breakable,colback=blue1,colframe=black]
\noindent
{\small 
\textbf{ALMI capability specification.} Two use cases (Figure~\ref{fig:almi}), defined in collaboration with potential users and their care takers at the ACE Alzheimer Center Barcelona (a major Spanish care provider for Alzheimer’s sufferers), were selected for prototyping in the ALMI project~\cite{stefanakos-2026}:
\begin{enumerate}
    \item Supporting a user with mild cognitive and physical impairments with the preparation of a simple pasta dish. This includes bringing the required ingredients to the user and providing reminders about the steps of the cooking process (boiling the water, adding the pasta to the pot with boiling water, etc.). Further to the functional capabilities needed for these tasks, subsequent research into normative requirements for the ALMI solution~\cite{feng2024analyzing} suggested the inclusion of the following additional capabilities derived from the principles of beneficence and human autonomy:
    \begin{itemize}
        \item \textbf{MealTime}---an event issued by a system timer informing the AI agent that the user’s scheduled meal time has been reached (critical for users with cognitive impairments); 
        \item \textbf{InformUser}---an event the agent can issue to invoke another component designed to notify the user about the meal time;
        \item \textbf{userOccupied}---a capability allowing the agent to check whether the user is engaged in personal activities that should not be interrupted;
        \item \textbf{RemindLater}---an event the agent can issue to delay notifying the user about the meal time.
    \end{itemize}
    \item Calling emergency support if the user is detected lying on the floor or in the case of a fire, requiring  high-level capabilities to spot a \textbf{HumanOnFloor}, to issue a \textbf{SmokeDetectorAlarm}, and to \textbf{CallEmergencySupport}, respectively. Awareness of the AI normative frameworks mentioned earlier suggested the further inclusion of a \textbf{humanAssents} capability enabling the agent to check if the user assents to support being called in the first of the two scenarios.
\end{enumerate}
Two types of capabilities are included in this (partial) list: instantaneous \emph{events} issued by the AI agent or other system components (capitalised, e.g., \textbf{MealTime}); and \emph{measures} that the agent reads as needed via a system API (starting with a lowercase letter, e.g., \textbf{userOccupied}).
} 
\end{tcolorbox}

\noindent
\textbf{Stage~2.~SLEEC requirements elicitation.} 
This stage translates abstract normative principles into actionable operational rules for the AI agent. This can be achieved, for instance, by using the approach from~\cite{townsend2022pluralistic}, which comprises a suite of activities engaging a representative selection of stakeholders, including ethicists, lawyers, sociologists, developers, regulators, and users: 
\begin{enumerate}
    \item \emph{Identification of high-level principles} (e.g., non-maleficence, human autonomy, and privacy) that must inform the agent design in light of its capabilities and  operational context. These principles are drawn from AI governance frameworks as well as ethical theories, codes of conduct, religious doctrines, customs, and social conventions, supplemented by input from domain experts, community groups, and the public.
    \item \emph{Derivation of normative-principle proxies}, i.e., of actionable placeholders that capture the value(s) of an abstract normative principle, and \emph{proxy mapping to agent capabilities} that can be used to achieve the proxy. For example, obtaining assent can act as a proxy for human autonomy, and can be mapped to the AI agent's voice-recognition capability `\textbf{humanAssents}'.
    \item \emph{Definition of SLEEC rules} by questioning the intention of the agent capabilities in light of their associated normative-principle proxies. This results in a set of normative rules specified in the SLEEC domain-specific language (DSL) from ~\cite{getir2023specification,yaman2023specification}, which was co-defined and validated with technical and non-technical stakeholders. The basic form of a SLEEC rule is:
    \begin{equation} \label{eq:rule}
  \text{\nterm{id} \hspace*{-2.5mm}
    \begin{array}[t]{l}
      \kw{when}~\nterm{triggerEvent}~[\kw{and} ~\nterm{triggerGuard}]~
      \kw{then}~\nterm{response}~[\kw{within}~\nterm{timeframe}],
    \end{array}  }
\end{equation}
where \nterm{id} is a unique rule identifier, \nterm{triggerEvent} is an 
event monitored by the agent, the optional \nterm{triggerGuard} is a Boolean expression over a set of \emph{measures} available to the agent, and \nterm{response} is an event defining the action that the agent shall perform (within the specified \nterm{timeframe}, if provided) when the \nterm{triggerEvent} occurs and, if specified, the \nterm{triggerGuard} is \smallsf{true}. These events, measures, and actions correspond to capabilities of the agent. One or more \emph{defeaters} of the general form below can be appended to a SLEEC rule~\eqref{eq:rule} to specify circumstances in which the rule does not apply in its base form: 
\begin{equation}
\label{eq:defeater}
  \text{
    \kw{unless}~\nterm{defeaterGuard}~[\kw{then}~\nterm{defeaterResponse}]}
\end{equation}
If at least one \nterm{defeaterGuard} Boolean expression of such a defeater is \smallsf{true}, then instead of responding as in the base rule, the agent shall perform the \nterm{defeaterResponse} corresponding to the outermost  \nterm{defeaterGuard}.  If that optional defeater response is not present or that outermost guard has no associated \nterm{defeaterResponse}, the rule does not apply at all.
\end{enumerate}
The successful completion of this stage yields a comprehensive, justified set of SLEEC requirements. We note that success may require returning to Stage~1 to revise the agent's capabilities by removing normatively problematic capabilities or adding missing capabilities required by SLEEC-rule triggers, defeaters, or responses. Failure to achieve a stable combination of capabilities and rules indicates that developing a SLEEC-compliant AI agent is infeasible. In such cases, halting the operationalisation process is a valid and necessary outcome, signaling that an agent meeting human expectations cannot be realised with currently available technologies or resources.

\begin{tcolorbox}[breakable,colback=blue1,colframe=black]
\noindent
{\small 
\textbf{ALMI SLEEC requirements elicitation.} The SLEEC rules in Table~\ref{tab:sleec_rules} illustrate the outcome of the Stage~2 elicitation process for a subset of ALMI’s core assistive tasks. Starting from high-level principles (i.e., beneficence, non-maleficence, and human autonomy), stakeholders identified actionable proxies (e.g., lack of human assent) and mapped them to the ALMI capabilities we described earlier, deriving SLEEC rules. For instance, rule R3 operationalises respect for autonomy in the context of contacting emergency support. Specifically, it constrains emergency escalation following a detected fall when the user does not assent, using lack of assent as a proxy for autonomy and specifying a temporary prohibition on contacting emergency support.

\vspace*{1mm}
\begin{center}
    \captionsetup{type=table} 
    \captionof{table}{\small Sample normative requirements for the ALMI assistive robot, expressed in SLEEC DSL.}
    \label{tab:sleec_rules}
    
    \vspace*{-2.5mm}
    \scalebox{0.9}{
        \begin{tabular}{l l}
        \toprule
             R1 & \kw{when} \textbf{MealTime} \kw{then} \textbf{InformUser} \kw{within} 10 \kw{minutes}
                 \kw{unless} \textbf{userOccupied} \kw{then} \textbf{RemindLater}  \\
             R2  & \kw{when} \textbf{SmokeDetectorAlarm} \kw{then} \textbf{CallEmergencySupport} \kw{within} 2 \kw{minutes} \\
             R3  & \kw{when} \textbf{HumanOnFloor} \kw{then} \textbf{CallEmergencySupport} \kw{within} 4 \kw{minutes} \\
             & \hspace*{8mm}\kw{unless}  (\kw{not} \textbf{humanAssents}) \kw{then} \kw{not} \textbf{CallEmergencySupport} \kw{within} 4 \kw{minutes}\\
        \bottomrule
        \end{tabular}
    }
\end{center}
} 
\end{tcolorbox}

\noindent
\textbf{Stage~3.~SLEEC requirements well-formedness checking.} This stage 
checks the SLEEC rule set from Stage~2 for  \emph{well-formedness issues} (WFIs). Such issues, often complex and subtle, arise from rules being defined by stakeholders with widely different expertise, e.g., ethicists, lawyers, regulators, and users. For a rigorous WFI analysis, rules must be given a formal semantics, allowing the use of established formal techniques to check their well-formedness. Any WFIs identified need to be ``debugged'' and resolved together with the relevant  stakeholders, which may necessitate returning to the elicitation stage. For normative requirements specified in the SLEEC DSL introduced earlier, well-formedness checking can be performed using two complementary tool-supported methods:
\begin{itemize}
    \item A process-algebraic method that captures the reactive behaviours (i.e., interactions, liveness, and timing) required or allowed by the SLEEC rules~\cite{getir2023specification} by mapping them to a timed variant of communicating sequential processes (CSP) process algebra called tock-CSP~\cite{BaxterRC22}. The analysis of the formalised rules with the FDR model checker~\cite{FDR} detects rule \emph{conflicts}~(pairs of SLEEC rules that cannot both always be obeyed for any implementation of the agent) and redundancies (pairs of rules for which obeying one rule always means that the other one is also obeyed). The steps of the approach are automated by the SLEEC-TK toolkit~\cite{yaman2024toolkit}, with WFI debugging and resolution supported by a complementary LLM-guided tool~\cite{kleijwegt2025tool}.
    \item A formal method~\cite{feng2024analyzing} that encodes a \emph{SLEEC ruleset as a whole} into first-order logic with relational objects~(FOL*)~\cite{10.1007/978-3-031-37709-9_18}, 
    enabling the use of a satisfiability checker (LEGOS) to analyse the \emph{global interactions of the ruleset}. The analysis produces, for several WFI types, causal (un)satisfiability proofs~\cite{feng-et-al-24c}, which are then translated into the SLEEC DSL for broad stakeholder accessibility.
    Beyond conflicts and redundancies, these include \emph{insufficiency} (ruleset failure to prevent normatively undesirable behaviours) and \emph{over-restrictiveness} (ruleset disallowing desirable behaviours). To enable this, the FOL* encoding explicitly incorporates \emph{concerns} (undesirable behaviours that the system must avoid) and domain \emph{facts} (core system functionalities that must remain achievable), ensuring that the ruleset is both sufficient to prevent harmful behaviours and not so restrictive as to undermine the system's intended purpose. The application of the approach is supported by the LEGOS-SLEEC tool~\cite{feng2024analyzing}. 
\end{itemize}
Overall, the process-algebraic approach targets pairwise well-formedness issues arising from the agent's sequential and timed execution of rules, while the logic-based LEGOS approach reasons over the ruleset as a whole to assess its overall  well-formedness, thereby providing complementary guarantees for the normative requirements. 

\begin{tcolorbox}[breakable,colback=blue1,colframe=black]
\noindent
{\small 
\textbf{ALMI SLEEC requirements checking.} As an illustration of well-formedness checking on the SLEEC rules elicited earlier for the ALMI robot, the methods above can be used to uncover different WFI types: 
\begin{enumerate}
    \item Using the process-algebraic approach, the joint analysis of rules R2 and R3 from Table~\ref{tab:sleec_rules} reveals a \emph{conflict}:~when the event \textbf{HumanOnFloor} occurs and the user does not assent to calling emergency services~(`\kw{not} \textbf{humanAssents}'), and the \textbf{SmokeDetectorAlarm} event is triggered within the next three minutes, rule R3 blocks the response required by rule R2 (i.e., calling emergency services within two minutes from the occurrence of  \textbf{SmokeDetectorAlarm}). One way of resolving this conflict is to reduce the length of time for which \textbf{CallEmergencySupport} events are prohibited in R3's defeater from 3~minutes to 1~minute. 
    \item Applying the LEGOS approach to the same rule set identifies an \emph{over-restrictiveness} WFI with respect to the system purpose. 
    Rule~R3 states that when \textbf{HumanOnFloor} occurs, emergency support must be called, \emph{unless} \textbf{not humanAssents}, in which case calling emergency support is prohibited. However, an unresponsive person is incapable of providing assent. Consequently, there exists no feasible trace in which the system calls emergency services for an unresponsive person on the floor, even though such intervention is required by the system’s safety purpose. 
    Resolving this WFI requires refining the defeater condition so that the prohibition of \textbf{CallEmergencySupport} applies only when both \textbf{not humanAssents} and a new measure, \textbf{userResponsive}, are satisfied. This ensures that emergency services can be summoned for an unresponsive fallen user, though it necessitates a new robot capability to detect user (un)responsiveness, e.g., by ``asking'' \emph{`Are you okay?'} and using deep-learning perception to check for a response. 
\end{enumerate}
By incorporating both fixes described above, the original rule R3 is refined into the rule R3$'$ from Table~\ref{tab:sleec_rules_revised}.

\vspace*{1mm}
\begin{center}
    \captionsetup{type=table} 
    \captionof{table}{\small Revised SLEEC rules for the ALMI assistive robot.}
    \label{tab:sleec_rules_revised}
    
    \vspace*{-2.5mm}
    \scalebox{0.9}{
        \begin{tabular}{l l}
        \toprule
             R1 & \kw{when} \textbf{MealTime} \kw{then} \textbf{InformUser} \kw{within} 10 \kw{minutes}
                 \kw{unless} \textbf{userOccupied} \kw{then} \textbf{RemindLater}  \\
             R2  & \kw{when} \textbf{SmokeDetectorAlarm} \kw{then} \textbf{CallEmergencySupport} \kw{within} 2 \kw{minutes} \\
             R3$'$  & \kw{when} \textbf{HumanOnFloor} \kw{then} \textbf{CallEmergencySupport} \kw{within} 4 \kw{minutes} \kw{unless}  ((\kw{not} \textbf{humanAssents}) \\
             & \hspace*{8mm}\kw{and}  \textbf{userResponsive}) \kw{then} \kw{not} \textbf{CallEmergencySupport} \kw{within} 1 \kw{minute}\\
        \bottomrule
        \end{tabular}
    }
\end{center}
} 
\end{tcolorbox}

\noindent
\textbf{Stage~4.~SLEEC-aware AI agent implementation.}
This stage integrates  the validated SLEEC requirements into the design, training, and deployment of the AI agent. The high-level agent capabilities identified in Stage 1 and refined through Stages 2 and 3 are mapped onto implementable agent abstractions such as observable events, monitored measures, internal state variables, and controllable actions. This refinement establishes the level of abstraction for SLEEC-rule enforcement in implementation, while preserving traceability to the validated normative requirements.

The refined rules are then incorporated into both \textcolor{blue}{training}-time exposure and runtime control. During training, dataset schemas are structured to explicitly encode situations that satisfy or violate SLEEC rules, allowing the agent to learn normatively relevant distinctions (e.g., when an action is permitted, prohibited, or requires additional checks). At deployment, the same rules are realised as runtime guardrails operating over the refined abstractions and co-deployed with the agent. These guardrails monitor the agent's perceptions and decisions, constrain its actions when rules apply, and provide explicit decision points for handling defeaters. Because these guardrails operate at a well-defined abstraction level, they can be revised, extended, or temporarily overridden in response to updated SLEEC requirements or contextual changes, providing a foundation for runtime adaptation and evolution of normative behaviour:~see the open challenges section of this paper.

\vspace*{-1mm}
\begin{tcolorbox}[breakable,colback=blue1,colframe=black]
\noindent
{\small 
\textbf{SLEEC-aware ALMI implementation.} To ensure SLEEC-awareness, the robot must enforce the rules from the previous stages---such as R1 (meal-time notification), R2 (calling emergency support after smoke alarms), and R3$'$ (restricted emergency calls after a user fall). This is achieved as follows:
\begin{itemize}
    \item The capabilities defined in Stage~1 (e.g., \textbf{MealTime}, \textbf{humanAssents}, and \textbf{CallEmergencySupport}) are mapped to implementable abstractions, including sensor-derived events (e.g., fall detection), dialogue outcomes (e.g., assent), and decision points within the agent's control logic. 
    \item The robot is trained on data explicitly encoding compliant and non-compliant situations, such as 
    falls with and without user assent, and meal-time notifications that are delivered or deferred. 
    These distinctions are encoded directly in the training data schema, rather than being left implicit.
    \item At runtime, SLEEC rules act as guardrails that continuously monitor these events. For instance, if \textbf{HumanOnFloor} is detected, a guardrail corresponding to rule R3$'$ prevents \textbf{CallEmergencySupport} unless \textbf{humanAssents} holds---except when a higher-priority trigger (e.g., a concurrent \textbf{SmokeDetectorAlarm}, rule R2) applies. As guardrails are decoupled from the core agent logic,     they can be updated     after consent or emergency-protocol changes without     full agent retraining.
\end{itemize}} 
\end{tcolorbox}

\noindent
\textbf{Stage~5.~Verification of AI agent compliance.} This stage focuses on formally verifying that the AI agent satisfies its SLEEC rule set. A conforming agent must monitor its environment to identify when each rule applies, accounting for rule triggers and defeaters, and responding as required by the rule. While conceptually simple, this is not catered for by standard notions of conformance, which cannot distinguish capabilities used to execute the agent's mission from those required to perceive the environment.
For example, a rule triggered when the temperature exceeds a threshold does not necessarily mandate the implementation of a temperature-reading mechanism if the agent can access that information by other means.  Indeed, there are many ways in which an agent may maintain knowledge of its environment, and the rules are not concerned with constraining them.

One approach for performing this verification is proposed in~\cite{getir2023specification}. This approach and its supporting tool~\cite{yaman2024toolkit} use a novel notion of conformance that extends the timewise refinement for tock-CSP~\cite{BaxterRC22} with a construct that reflects the special nature of the measures from SLEEC rules. This enables the use of model checking or theorem proving to verify if a CSP-based model of an AI-agent conforms to a set of SLEEC rules by leveraging agent models written in RoboChart~\cite{MRLCTW19}, a diagrammatic domain-specific modelling language with a tock-CSP semantics. Given a RoboChart model, its semantics can be automatically generated by the RoboChart tool~\cite{robotool}, and used for automated verification of conformance to SLEEC rules, or the provision of counterexamples when rules are violated. Importantly, once a RoboChart model is verified as SLEEC-compliant, existing tools~\cite{Cavalcanti2021} can automate its translation into code for simulation or deployment, ensuring that the formal normative guarantees established during verification are reflected in the operational agent.

\begin{tcolorbox}[breakable,colback=blue1,colframe=black]
\noindent
{\small 
\textbf{ALMI SLEEC-compliance verification.} Consider a RoboChart model for a candidate ALMI design in which its engineering team inadvertently prioritised the robot finishing its current task before switching to emergency protocols. As a result, when the robot is fetching meal ingredients, its typically immediate response to a \textbf{HumanOnFloor} event is delayed. Given this RoboChart model and rule R3, the automated verification tool~\cite{yaman2024toolkit} identifies a violation and generates the counterexample trace:

\begin{list}{}{\leftmargin=5mm \rightmargin=5mm}
\item
{\footnotesize
\textbf{MealTime}, \textbf{userOccupied}.false, \textbf{InformUser}, tock, \textbf{FetchingIngredients}, \textbf{HumanOnFloor}, tock, tock, tock, \textbf{AbandonFetchingIngredients}, tock, tock, \textbf{humanAssents}.true, \textbf{CallEmergencySupport}
}    
\end{list}

This trace illustrates a scenario in which rule R3 is violated because more than the allowed four minutes (represented by five instances of the discrete time counter `tock') elapse between the trigger event \textbf{HumanOnFloor} and the required response \textbf{CallEmergencySupport}. Presented with this counterexample, the engineering team must revise the design to ensure that abandoning a non-critical ongoing task does not impede a timely response to a user fall.
} 
\end{tcolorbox}

\noindent
\textbf{Process output: deployed/cancelled AI agent.} The SLEEC  process serves a dual purpose: to facilitate AI adoption where it can provide meaningful functionality in a verifiable, normatively compliant manner, and to prevent the development and deployment of AI agents that can violate human norms. Cancellation occurs when any stage of the process cannot be completed successfully, not even through revisiting earlier stages. This may arise, for instance, from stakeholder disagreement on requirements, technical or financial infeasibility discovered as the process identifies the need for additional agent capabilities, or a loss of utility after  planned capabilities are removed to ensure normative compliance. We note that the cancellation of a project on any of these grounds represents a positive outcome, which not only preempts deploying an AI agent that can violate norms, but may also identify gaps requiring further regulatory and technical efforts.

\section{Open Challenges}

Significant challenges remain to be addressed to ensure the effective development of normatively compliant AI agents. We outline these challenges and suggest avenues for addressing them below.

\smallskip\noindent
\textbf{Normative-principle reification} for concrete AI agents poses a significant challenge to their developers, regulators, and other stakeholders. While international efforts have established broad classes of normative values and principles~\cite{BS-2022,ieee-2021,iso-iec-2022,oecd-2024,unesco-2021}, their abstract nature often complicates the mapping process to concrete technical specifications. For instance, the UNESCO recommendations on the ethics of AI~\cite{unesco-2021} identify values such as \emph{`respect, protection and promotion of human rights and fundamental freedoms and human dignity'} and principles like \emph{`human oversight and determination'}. Translating these high-level abstractions into practical SLEEC requirements for an assistive dressing robot, a self-driving car, or a medical diagnosis system is a substantial, application-specific undertaking. Although domain-specific guidance like the WHO’s ethics and governance guidance for large models in healthcare~\cite{WHO-2025} is emerging, these efforts remain limited. Expanding such sector-specific regulatory frameworks, alongside the creation of shared normative-requirement repositories for specific classes of applications, could mitigate the complexity of this hurdle.

\smallskip\noindent
\textbf{Normative ambiguity and value-conflict resolution} present further conceptual and practical challenges in the operationalisation of AI norms. Normative values are inherently multifaceted, so that there can be different interpretations by stakeholders, potentially precluding a consensus on the specific requirements for an AI agent. Furthermore, alignment with all relevant principles is typically unattainable, as values may conflict in practice; for instance, requirements for transparency may, in certain contexts, impede user privacy, or safety constraints may limit personal autonomy. Finally, as Alan Turing famously noted, \emph{`it is not possible to produce a set of rules purporting to describe what a [person] should do in every conceivable set of circumstances'}~\cite{turing1950}. This highlights the difficulty of determining when a SLEEC ruleset---and the defeaters underpinning its logic---can be considered ``sufficient''. Addressing these ``normative frictions'' requires not only technical mechanisms for managing priority-based trade-offs (e.g., through systematically identified defeaters~\cite{townsend2025normative}), but also the development of  deliberative methods (and formal stopping criteria) enabling stakeholders to negotiate and settle on agreed value hierarchies appropriate for an AI agent under development.

\smallskip\noindent
\textbf{SLEEC-capable agent engineering} involves technical hurdles including \emph{traceability}, \emph{timing}, and \emph{computational capacity}. First, faithfully implementing abstractly described capabilities creates a \emph{traceability} challenge, as compliance verification requires a mapping between high-level normative concepts and low-level agent actions.
Second, supporting the SLEEC notion of \emph{time} poses an engineering difficulty. While normative concepts such as user consent are ongoing processes only requiring renewal over days or weeks, the sensor processing and decision-making cycles of AI agents operate in fractions of a second. Reconciling these very different temporal scales is essential for the agent to maintain a consistent model of the environment in which it operates, and to understand the frequencies at which various normative measures must be monitored and refreshed.
Finally, the \emph{computational capacity} required to implement advanced human-agent interaction capabilities, and to distinguish between subtly different SLEEC contexts can be prohibitive. Accurately inferring environmental status and user intent is resource-intensive, often requiring advanced hardware and software architectures. 
For these reasons, developing an agent that verifiably satisfies a SLEEC ruleset is inherently challenging. The use of digital twins for high-fidelity simulation of the agent and its environment could provide a way of mitigating this, but further research is required to formally link such simulations to the SLEEC requirements the agent must adhere to.

\smallskip\noindent
\textbf{Runtime adaptation to changing normative requirements}, essential to ensure AI agent alignment with the SLEEC norms of users and contexts not encountered before, poses further technical challenges. While the process we proposed facilitates the operationalisation of SLEEC requirements at development time, the specific social or cultural preferences of an individual user, or the nuances of a particular operational setting, may remain unknown until deployment. Humans adapt to such variations seamlessly, e.g., by modifying their provision of a service to the norms and values of its beneficiary. Equipping AI agents with similar flexibility, however, remains a complex undertaking. As explored in recent work on norm-aware workflow adaptation~\cite{sleec-adapt-2025}, addressing this requires moving beyond static SLEEC rulesets towards agents capable of dynamically updating their behaviour in response to runtime normative signals---and of perceiving these signals in the first instance. Developing the mechanisms required to achieve such normative adaptivity, while maintaining the formal guarantees established during the development-time operationalisation process, is essential for the deployment of agents in open, human-centric environments.

\smallskip\noindent
\textbf{Normative and AI competence}, required to meaningfully contribute to the operationalisation of SLEEC norms, is difficult to develop and maintain across the diverse stakeholder groups involved in AI solution development and operation. If AI agents are to align with complex societal expectations, the professional preparation of all participants---not only engineers but also regulators, legal experts, and domain specialists---must be reconsidered. This is needed to overcome the current lack of a shared understanding and common language, which often leads non-technical stakeholders to propose normative requirements that are technically infeasible, and engineers to overlook the subtle nuances of SLEEC requirements. Addressing this requires a shift towards multidisciplinary education that provides technical professionals with a broad understanding of SLEEC principles as core engineering considerations, and equipping non-technical professionals with a foundational grasp of AI capabilities and limitations. We therefore call on academic institutions and professional societies to recognise normative and AI competence as foundational skills, and to support curricular frameworks, shared educational resources, and community standards that prepare multidisciplinary teams of graduates to co-create AI systems that are responsible, context-aware, and worthy of trust.

\section{Conclusion}

As AI agents are increasingly entrusted with high-stakes decisions, their alignment with social, legal, ethical, empathetic, and cultural (SLEEC) norms has become a primary engineering concern. Our proposed SLEEC-norm operationalisation process addresses the major limitations of traditional requirements engineering approaches, providing a rigorous methodology for refining abstract normative principles into verifiable technical requirements. While many activities within this process are supported by recently introduced SLEEC-norm elicitation, formalisation, validation, and verification methods, the progress towards fully norm-aware AI agents remains contingent on addressing significant normative, technical, and stakeholder competency challenges. Thus, our SLEEC-norm operationalisation process also outlines a critical research and policy agenda for achieving such awareness of human norms and values within AI agents.

\section*{Acknowledgements}

This project has received funding from the UKRI project EP/V026747/1 `Trustworthy Autonomous Systems Node in Resilience', the RAi UK project `Disruption Mitigation for Responsible AI',  IVADO Distinction Starting Grant, NSERC-CSE grant `An End-to-End Approach to Safe and Secure AI Systems', and Royal Academy of Engineering grant CiET1718/45.

\end{document}